\documentclass[journal,twoside,web]{ieeecolor}
\usepackage{jsen}
\usepackage{cite}
\usepackage{amsmath,amssymb,amsfonts}
\usepackage{algorithmic}
\usepackage{graphicx}
\usepackage{bm}
\usepackage{multirow}
\usepackage{textcomp}
\usepackage{threeparttable}
\usepackage{makecell}
\usepackage{verbatim}
\usepackage{url}
\usepackage{color}
\usepackage{xcolor}
\usepackage{booktabs}
\usepackage{wrapfig}
\def\BibTeX{{\rm B\kern-.05em{\sc i\kern-.025em b}\kern-.08em
    T\kern-.1667em\lower.7ex\hbox{E}\kern-.125emX}}
\definecolor{abstractbg}{rgb}{0.89804,0.94510,0.83137}
\begin{document}
	
	 % insert the first page about the copyright of IEEE
	\twocolumn[
	
	\begin{center}
		This paper has been accepted for publication in \emph{IEEE Sensors   Journal}. \vspace*{13pt}
		
		% DOI: {\color{blue}
		% \href{http://dx.doi.org/10.1109/TITS.2019.2961705}
		% {10.1109/TITS.2019.2961705}}
		
		\vspace*{23pt}
	\end{center}
	
	\hspace{0.5em}\copyright 2023 IEEE. Personal use of this material is permitted. Permission from IEEE must be  obtained for all other uses, in any  current or future media, including reprinting/republishing this material for advertising or promotional purposes, creating new collective works, for resale or redistribution to servers or lists, or reuse of any copyrighted component of this work in other works.
	]
	% new page
	\clearpage

	\thispagestyle{empty}
	\pagestyle{empty}

\title{PVI-DSO: Leveraging Planar Regularities for  Direct Sparse Visual-Inertial Odometry}
\author{Bo Xu, Xin Li, Jingrong Wang, Chau Yuen, \IEEEmembership{Fellow, IEEE}, Jiancheng Li
% \thanks{This paragraph of the first footnote will contain the date on 
% which you submitted your paper for review. It will also contain support 
% information, including sponsor and financial support acknowledgment. For 
% example, ``This work was supported in part by the U.S. Department of 
% Commerce under Grant BS123456.'' }
\thanks{Bo Xu, Jiancheng Li are with School of Geodesy and Geomatics, Wuhan University, Wuhan 430079, China; Corresponding author: Jiancheng Li, Email: jcli@whu.edu.cn
}
\thanks{Xin Li, Chau Yuen are with the Engineering Product Development at the Singapore University of Technology and Design (SUTD), 8 Somapah Road, 487372, Singapore;
% (Email: yuenchau@sutd.edu.sg).
}
\thanks{Jingrong Wang is with the GNSS research center, Wuhan University, Wuhan 430079, China;}}

\IEEEtitleabstractindextext{%
\fcolorbox{abstractbg}{abstractbg}{%
\begin{minipage}{\textwidth}%
\begin{wrapfigure}[12]{r}{3in}%
\includegraphics[width=3in]{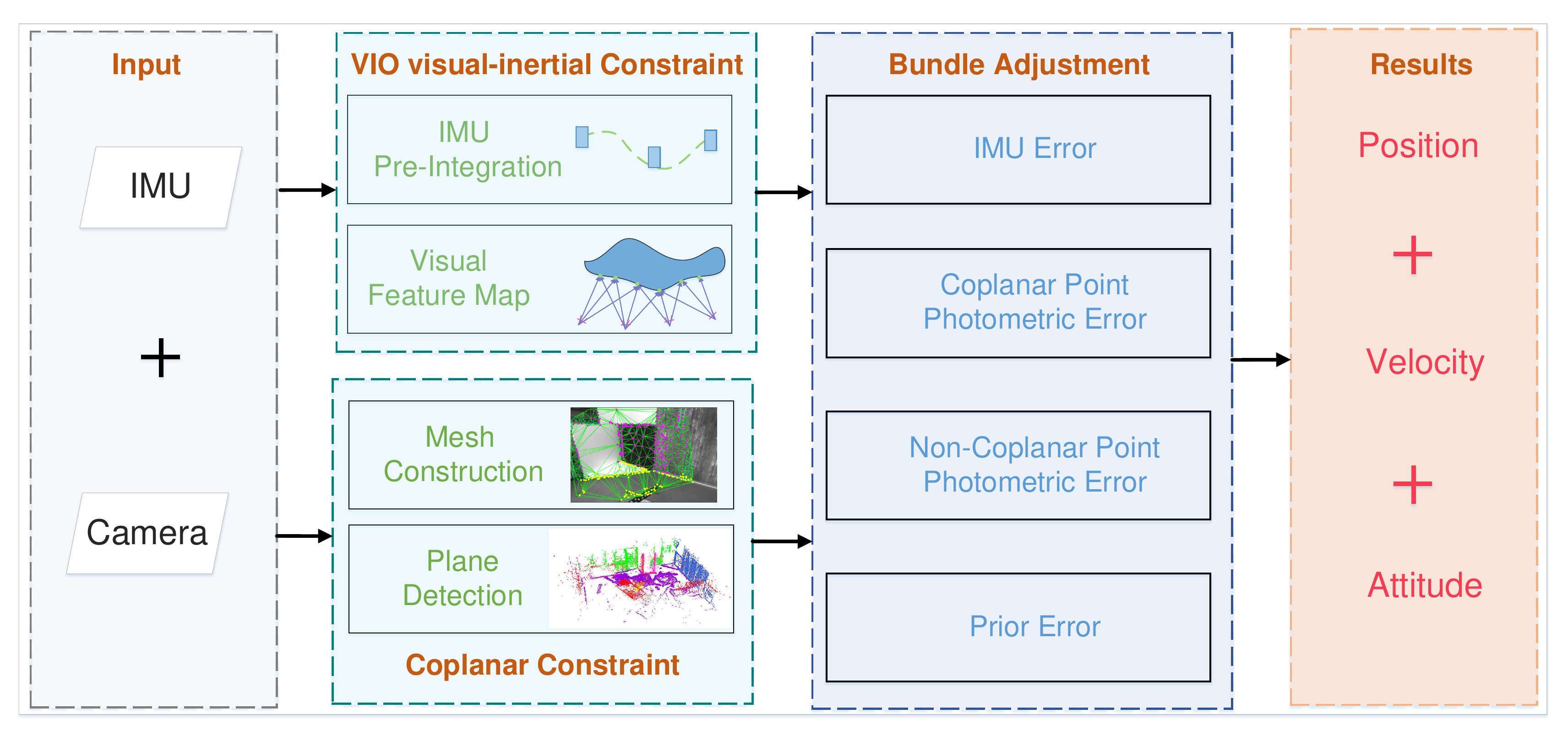}%
\end{wrapfigure}%
\begin{abstract}
The monocular visual-inertial odometry (VIO) based on the direct method can leverage all available pixels in the image to simultaneously estimate the camera motion and reconstruct the denser map of the scene in real time. However, the direct method is sensitive to photometric changes, which can be compensated by introducing geometric information in the environment.
In this paper, we propose a monocular direct sparse visual-inertial odometry, which exploits the planar regularities (PVI-DSO). 
Our system detects the planar regularities from the 3D mesh built on the estimated map points. To improve the pose estimation accuracy with the geometric information, a tightly coupled coplanar constraint expression is used to express photometric error in the direct method. Additionally, to improve the optimization efficiency, we elaborately derive the analytical Jacobian of the linearization form for the coplanar constraint. Finally, the inertial measurement error, coplanar point photometric error, non-coplanar photometric error, and prior error are added into the optimizer, which simultaneously improves the pose estimation  accuracy and mesh itself. We verified the performance of the whole system on simulation and real-world datasets. Extensive experiments have demonstrated that our system outperforms the state-of-the-art counterparts.

\end{abstract}

\end{minipage}}}

\maketitle

\section{Introduction}
\label{sec:introduction}
\IEEEPARstart{S}{imultaneous} localization and mapping (SLAM) are research hotspots in the field of robots, autonomous driving, augmented reality, etc. Camera and inertial measurement units (IMU) are low-cost and effective sensors. Visual inertial odometry (VIO) combines the complementary of the two sensors to improve the accuracy and robustness of the pose estimation. Existing VIO methods \cite{VINS, ORBSLAM3, mourikis2007multi} generally form visual observations based on the feature point (indirect) method. However, in the weak texture environment, the lack of effective point features extracted by the indirect method makes the system fail to estimate the poses. Fortunately, the direct method \cite{engel2017direct, von2018direct} can utilize all available pixels of images to generate a more complete model, which effectively improves the performance of VIO in  weak texture environments. However, the sensitivity to photometric changes makes it difficult to accurately estimate the depth of pixels.

It has been shown that the geometric features (e.g., lines and planes) in the environment can provide valuable information to VIO, and introducing  additional constraints guides the optimization processes in the VIO system. Line features are already commonly used in  SLAM systems \cite{PL-VIO,pumarola2017pl,structslam, li2018direct}. Compared with straight lines, plane features  cannot be accurately recognized, which makes it difficult to introduce plane constraints into SLAM. The learning-based methods have made significant progress in plane detection in recent years \cite{yu2019single, liu2019planercnn}. Nevertheless, these algorithms rely on GPUs, which consume relatively high amounts of power, making them impractical for computationally-constrained systems. Building lightweight 3D meshes based on the real-time map generated by VIO and  extracting the scene structure from the meshes become a possible way of plane detection \cite{rosinol2019incremental}. However this method needs the system to produce denser maps efficiently. However, the indirect VIO only provides a sparse map of 3D points \cite{wu2022enforcing, li2020leveraging}, which is hard to identify planar features. The dense mapping methods \cite{newcombe2010live, newcombe2011dtam} based on monocular vision and pixel-wise reconstruction lead to high time complexity and loss accuracy due to decoupling the trajectory estimation and mapping.

\begin{figure}[hptb]
	\centering
	\includegraphics[width=\columnwidth]{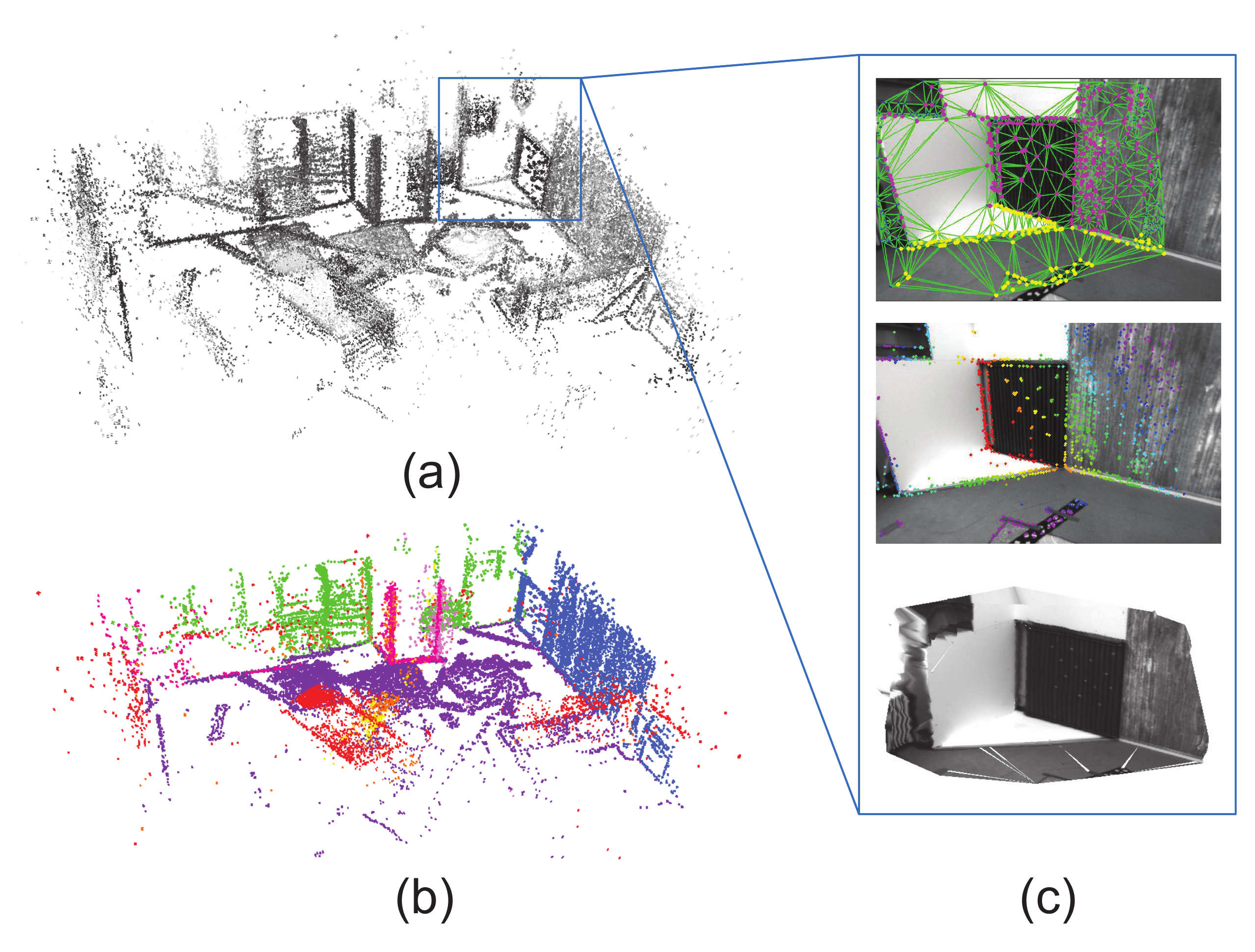}
	\caption{The proposed direct sparse visual-inertial odometry builds a denser map running on the V11 sequence of the  EuRoC dataset. (a) is the reconstruction map of the whole scenes. (b) is the coplanar points on different planes, which are distinguished by different colors. (c) is the 2D Delaunay triangulation generated in the depth map, raw depth map, and the reconstructed 3D mesh of the corner in the scene.}
	\label{fig:cover_grap}
\end{figure}

For the visual odometry (VO) / VIO based on the direct method, the accuracy of the pose estimation can be further improved after introducing the planar information. Wu and Beltrame \cite{wu2021direct} fuse the coplanar constraints extracted from color image using PlaneNet \cite{liu2018planenet} into DSO \cite{engel2017direct}. However, the CNN-based technique is time-consuming and can only be applied to global shutter color images. Concha and Civera \cite{concha2015dpptam} model the environment as high-gradient and low-gradient areas based on LSD-SLAM \cite{engel2014lsd}. The plane features are segmented from low-gradient areas with superpixels \cite{achanta2012slic} and estimated using Singular Value Decomposition (SVD) for the points clustered in each superpixel. However, in texture-rich scenes, the assumption of extracting the plane information from low-gradient areas is not satisfied, resulting in wrong plane segmentation. We noticed that the visual module in the direct method tracks the pixels with large enough intensity gradients. As shown in Fig.\ref{fig:cover_grap}, sufficient visual observation makes the reconstructed map denser. Therefore,
it is easy to extract plane regularities from 3D mesh built on a real-time estimated denser map, which brings almost negligible computation burden. Furthermore, introducing geometric information that is less sensitive to the photometric changes into VIO can benefit both state estimation and mapping.

This paper proposes a direct sparse visual-inertial odometry that leverages planar regularities, called PVI-DSO, which is an extension of DSO \cite{engel2017direct}. 
To improve the accuracy and robustness of the system, IMU measurements and coplanar information are integrated into the system. We detect the probable planar regularities from the 3D mesh generated by the real-time estimated point cloud. Then inspired by the method in \cite{li2020co}, a tightly coupled coplanar constraint expression is used to construct photometric error. Finally, we derive the analytic Jacobian for the linearized form of the coplanar constraint and present it in the Appendix. In summary, the main contribution of this work include: 

\begin{itemize}
\item We introduce the planar regularities which are detected through the 3D mesh in the direct method based VIO to improve the accuracy of the system. The 3D mesh segmentation which is performed on the real-time estimated denser map not only couples the state estimation and mapping, but also brings marginal computation burden.

\item  We adopt a tightly coupled coplanar constraint expression  in the direct method to construct the photometric error. For the efficiency of the optimization, the analytical Jacobian in the linearization form for the coplanar constraint is derived in detail.   

\item We design  extensive experiments on the simulation data, the challenging EuRoC dataset \cite{Burri2016}, and TUM VI dataset \cite{Schubert2018}. Experimental results demonstrate our system outperforms the state-of-the-art methods in  pose estimation.
\end{itemize}

\section{RELATED WORK}\label{sec:related work}
The most common features used in the SLAM / VIO algorithms are the points \cite{mourikis2007multi, VINS, ORBSLAM3, engel2017direct, forster2014svo}. As a complement to the point features, the geometric information (line, plane) introduced in the system with point features can improve the accuracy of the pose estimation and mapping, which have received extensive attention in recent years.

\textbf{Indirect method with geometric Regularities} \ Since the line features exist widely in the environment, it seems natural to fuse the line features into the framework based on points \cite{PL-VIO,pumarola2017pl, Tsai2018}. Furthermore, in the structural scenario, the lines with three perpendicular directions of the Manhattan world model can encode the global orientations of the local environments, which are utilized to improve the robustness and accuracy of pose estimation \cite{Zou2019, structslam, xu2021leveraging}. For the planar information, the main difficulty is how to accurately extract the planar regularities in the environment. Some works \cite{salas2014dense,zhang2019point,ma2016cpa} extract plane features with the assistance of depth maps obtained by the RGBD camera. Rosinol et al. \cite{rosinol2019incremental} propose a stereo VIO fusing plane information, called Mesh-VIO. Mesh-VIO extracts the planar regularities from the 3D meshes, which are generated by projecting 2D Delaunay triangulation based on 2D points to corresponding 3D points. However, sparse point clouds generated by indirect-based vision algorithms may result in an inaccurate 3D mesh reconstruction. More recently, Li et al. \cite{li2019robust} propose PVIO which leverages multi-plane priors in the VIO. The system uses the 3-point RANSAC method to fit the plane among the estimated 3D points with the RGB camera. Nevertheless, the RANSAC-based method is not stable when there are multi potential planes in the environment.

\textbf{Direct method with geometric Regularities} \ The most prominent direct VO approach in recent years is direct sparse odometry (DSO). To improve the stability of DSO, the IMU measurements are fused into DSO, including: VI-DSO \cite{von2018direct} with dynamic marginalization and  DM-VIO \cite{von2022dm} with delayed marginalization. Meanwhile, structural information in the environment  provides additional visual constraints, which can be used to reduce the drift of the estimator. For the line features utilized in the direct method, some works \cite{yang2017direct, li2018direct, gomez2016pl} force the 3D points in the map that satisfy the collinear constraints using straight lines, but not jointly optimizing the estimated poses. Zhou et al. \cite{zhou2021dplvo} introduce the collinear constraints into the DSO more elegantly. The 3D lines, points, and poses within a sliding window are jointly optimized. Cheng et al. \cite{cheng2020direct} extract the Manhattan world regularity from the line features in the image and merges the structural information into the optimization framework of photometric error. Currently, there are few works on fusing planar features in the direct method. As mentioned previously, the CNN-based plane detection method  \cite{wu2021direct} consumes a lot of computing resources. The superpixel segmentation method \cite{concha2015dpptam} leads to  wrong segmentation of planes in texture-rich scenes.

 \begin{figure}[hptb]
	    \centering
	    \includegraphics[width=\columnwidth]{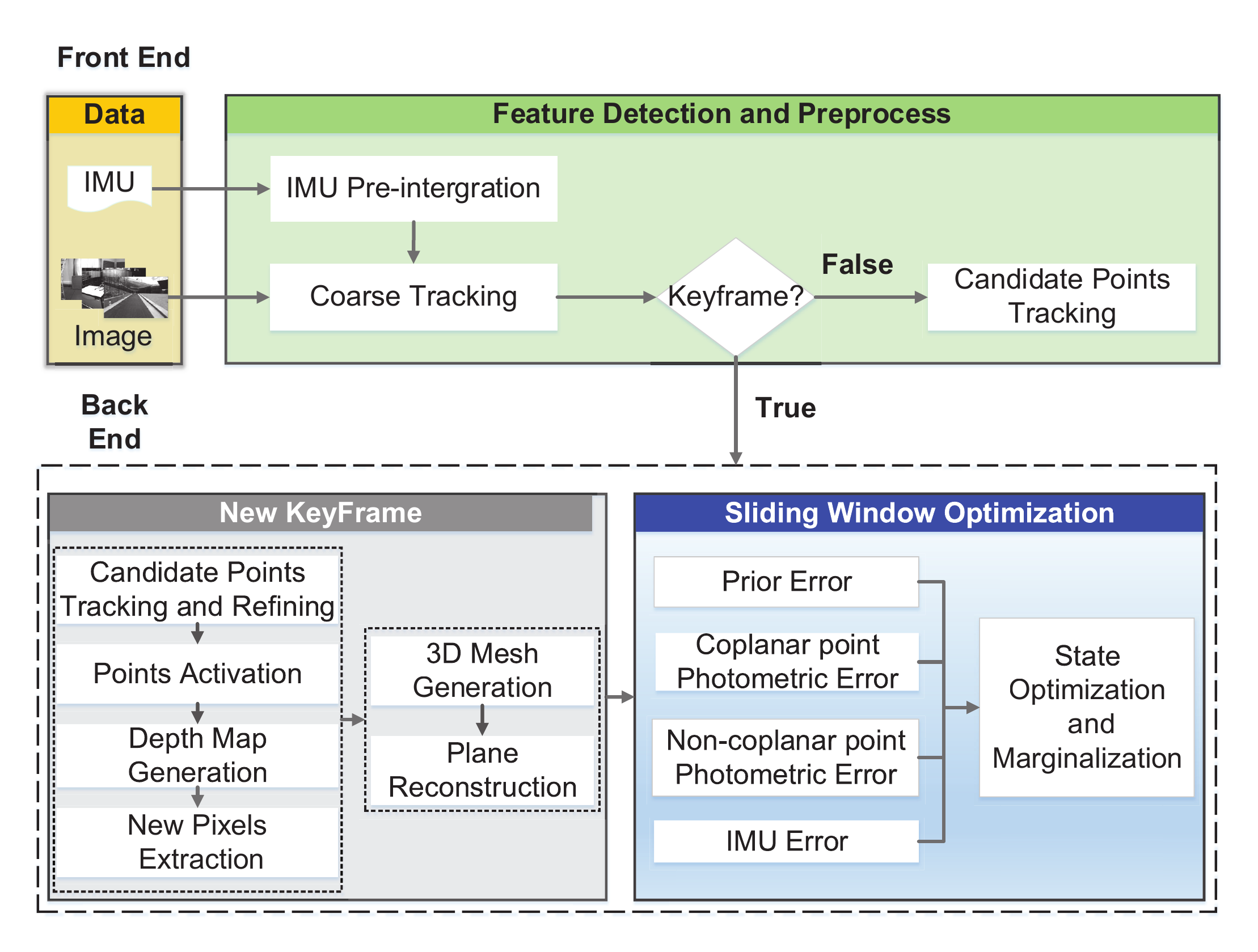}
	    \caption{Overview of our PVI-DSO system.}	\label{fig:pipeline}
\end{figure}
\section{SYSTEM OVERVIEW}\label{sec:system overview}

The system proposed in this paper is based on DSO\cite{engel2017direct}. We introduce the IMU measurements and geometric information in the environment to improve the accuracy and robustness of the system. As shown in Fig. \ref{fig:pipeline}, the proposed system contains two main modules running in parallel, the front end and the back end.
	
In the front end, we perform the coarse tracking to obtain the pose estimation and the photometric parameters estimation of the current frame, which serves as the initialization for the optimization in the back end.  With the aid of IMU integration, the coarse tracking is executed for every frame, in which the direct image alignment based on the photometric bundle adjustment is used to estimate the state variables of the most recent frame. And then, we determine whether the current frame is the keyframe through the criterion described in \cite{engel2017direct} . If the current frame is a keyframe, it will be delivered to the back end module. Otherwise, it is only used to track the candidate points to update the depth. To do so, we search along the epipolar line to find the correspondence with the minimum photometric error.

In the back end, the candidate points are tracked and refined with the latest keyframe to obtain more accurate depth. To activate the candidate points with low uncertainty, all active points in the sliding window are projected onto the most recent keyframe. The candidate points (also projected into this keyframe) are activated as active points when their distance to any existing point is maximum \cite{engel2017direct}. Then, the planar regularities are extracted from the activated points in the point clond.  We construct the 3D meshes through the depth map which is generated by the active points. The coplanar constraints and non-coplanar constraints are obtained through the segmentation of the 3D meshes. Finally, the operations of the visual-inertial bundle adjustment are executed. We add the non-coplanar point residuals, coplanar point residuals, inertial residuals, and corresponding prior residuals into the optimizer, which simultaneously improves the accuracy of the pose estimation and the mesh itself.

\section{Notations and Preliminaries} \label{sec:algorithm}
	
In this section, coordinate transformations are defined and the representations of point and plane features are given. Throughout the paper,
we denote vectors as bold lowercase letters $\mathbf{x}$, matrices as bold uppercase letters $\mathbf{H}$, scalars as lowercase letters $\lambda$, and functions as uppercase letters $E$. 

\subsection{Coordinate Transformation} \label{alg: coordinate system}
The world coordinate system is defined as a fixed inertial framework in which the $z$-axis is aligned with the gravity direction. $\mathbf{T}_{wi} \in \mathbf{SE}(3)$ represents the transformation from the IMU frame to the world frame, where $\mathbf{R}_{wi} \in \mathbf{SO}(3)$ and $\mathbf{t}_{wi} \in \mathbb{R}^3$ are the  rotation and translation, respectively. The transformation from the camera frame to the IMU frame is defined as $\mathbf{T}_{ic} \in \mathbf{SE}(3)$, and the transformation from the camera frame to the world frame $\mathbf{T}_{wc}$ can be calculated by: $\mathbf{T}_{wc} = \mathbf{T}_{wi}\mathbf{T}_{ic}$. Similarly, $\mathbf{R}_{ic}$ and $\mathbf{R}_{wc}$ represent the rotations from the camera frame to the IMU frame and from the camera frame to the world frame.  $\mathbf{t}_{ic}$ and $\mathbf{t}_{wc}$ are the corresponding translations. 

\begin{figure}[t]
	\centering
	\includegraphics[width=0.8\linewidth]{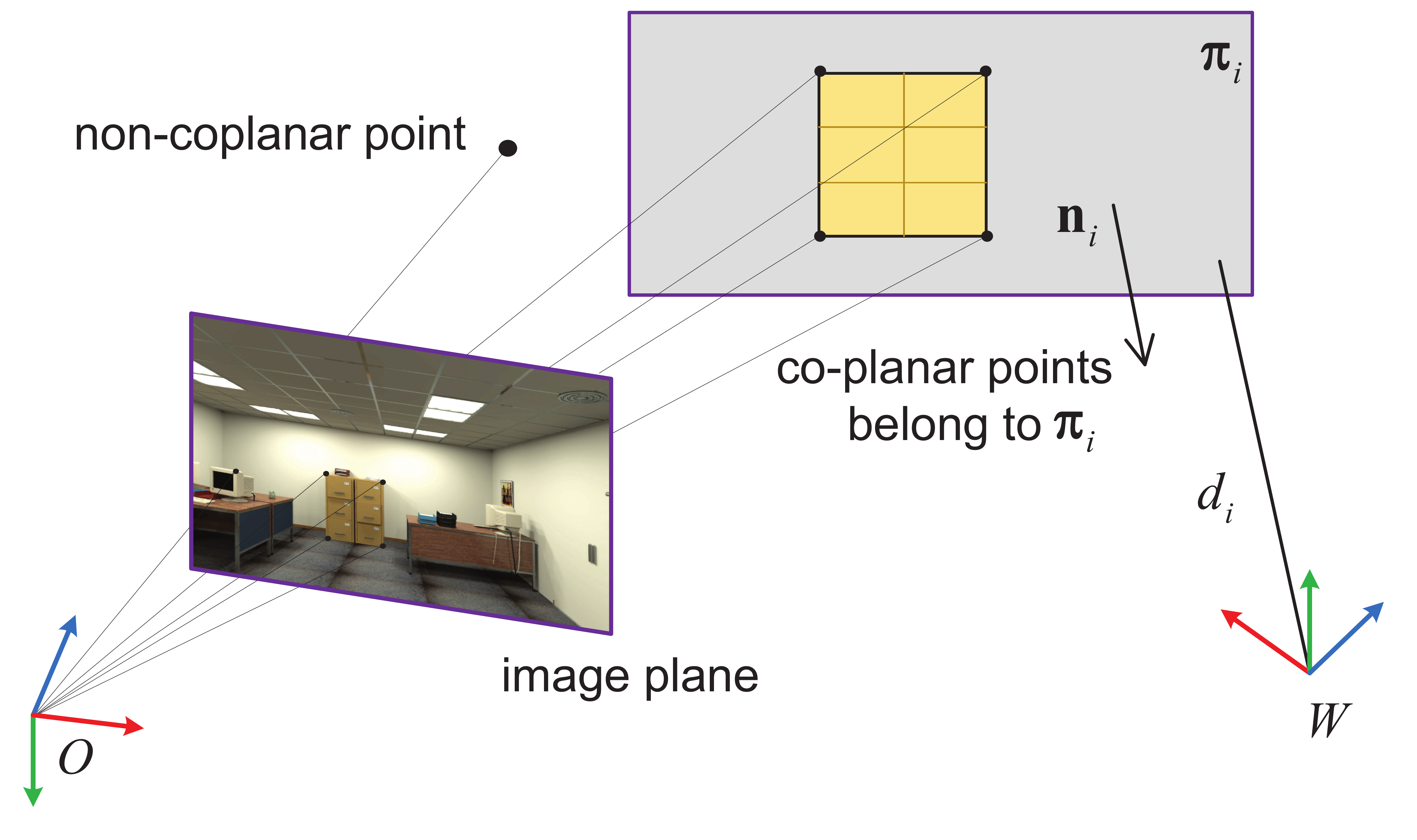}
	\caption{The coplanar points in the image lie on the plane $\bm{\pi}_i$ in the world frame, which is represented by a normal vector $\mathbf{n}_i$ and a distance $ \boldmath{d}_i$. }
	\label{fig:point_planar_nonplanar}
\end{figure}
\subsection{Point Representation} 
We use the inverse depth $d_p\in \mathbb{R}$ to parameterize the pixels from the host image frame in which it is extracted. With this parametric expression, we can project the pixels in the host image frame $I_h$ into the target image frame $I_t$ where it is observed. Assuming $\mathbf{p}_i$ is the pixel in the $I_h$,  the projection point $\mathbf{p}_i'$ in the $I_t$ is given by: 
\begin{flalign} 
	\begin{array}{c}
	\mathbf{p}_i' = \Pi_c\left(\mathbf{R}_{c_tc_h}\Pi_c^{-1}\left(\mathbf{p}_i, d_p\right) + \mathbf{t}_{c_tc_h}\right)
	\end{array}\label{fla: point transform}
\end{flalign} 
where $\mathbf{R}_{c_tc_h}$ and $\mathbf{t}_{c_tc_h}$ are the relative rotation and translation from image frame $I_h$ to $I_t$, $\Pi_c$ and $\Pi_c^{-1}$ are the projecton and back projection of the camera.
\subsection{Plane Representation}
As is shown in Fig.\ref{fig:point_planar_nonplanar}, a plane in the world frame can be represented by the Hessian normal form $\bm{\pi}_w =\begin{bmatrix}\bm{n}_w, d_w \end{bmatrix}^{\rm T}$, where $\bm{n}_w =\begin{bmatrix}n_x,n_y, n_z\end{bmatrix}$ is the normal of the plane, representing its orientation and $d_w$ is the distance from the origin of the world frame to the plane. The normal vector $\bm{n}_w$ has three parameters but only two Degrees of Freedom (DoF) with $||\bm{n}_w||_2 = 1$. 

\begin{figure}[t]
	\centering
	\includegraphics[width=0.55\linewidth]{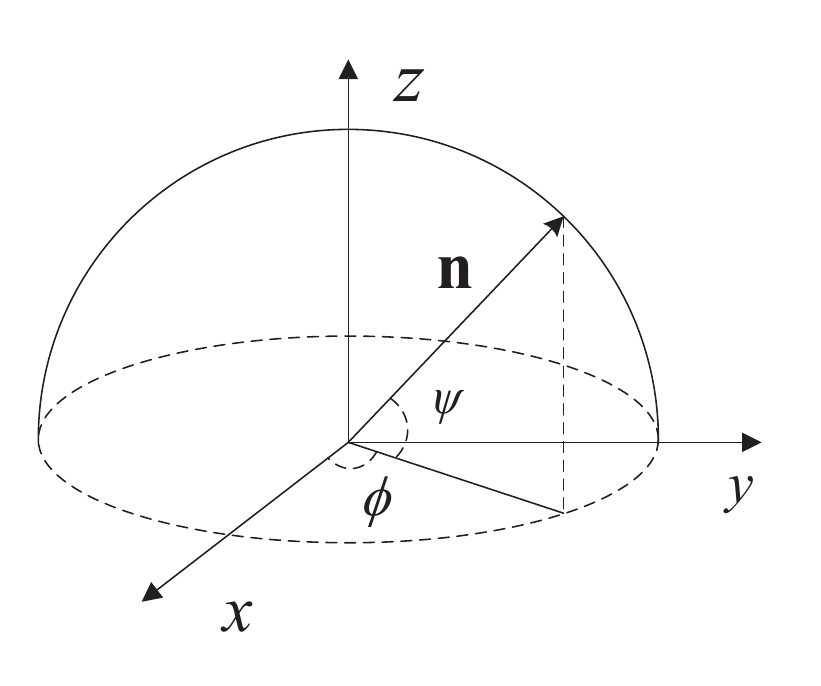}
	\caption{ 2-DoF parameter representation of the normal vector $\mathbf{n}$.}
	\label{fig:plane_parameter}
\end{figure}

To get a minimal parameterization of $\bm{\pi}_w$ for optimization, we represent it as $q(\bm{\pi}_w) = \begin{bmatrix}\phi, \psi, d_w\end{bmatrix}$, where $\phi$ and $\psi$ are the azimuth and elevation angles of the normal vector and $d_w$ is the distance from the Hessian form. As shown in Fig. \ref{fig:plane_parameter}, $\bm{\pi}_w$ is given by: 
 \begin{flalign} 
	\bm{\pi}_w= \begin{bmatrix} \cos(\psi)\cos(\phi), \cos(\psi)\sin(\phi), \sin(\psi) , d_w\end{bmatrix}^{\rm T}
\end{flalign}

Although we express the plane parameters in the world frame, we need to transfer the plane parameters to the camera frame to build the photometric error as described below. The transformation from world frame to camera frame is given by:  
\begin{flalign} 
	\bm{\pi}{_c}= \textbf{T}_{cw}^{-\rm T}\bm{\pi}{_w}
	\label{fla: transformation matrix of plane}
\end{flalign}
where $\mathbf{T}_{cw}$ is the transformation from the world frame to the camera frame. 
\section{VIO WITH COPLANAR REGULARITIES } \label{sec: direct sparse visual-inertial odometry}
In this section, the mechanism of coplanarity detection based on the direct method is first described. Then, the residuals of non-coplanar points, coplanar points, inertial measurements, and prior are emphasized. All the state variables in the sliding window are estimated by minimizing the sum of the energy function from visual residuals, IMU residuals, and prior residuals:
\begin{flalign} 
E_{total} = \lambda \cdot \left( E_{point} +  E_{point}'\right) + E_{inertial} \label{fla: total cost function} + E_{prior}
\end{flalign}
where $E_{point}$, $E_{point}'$, $E_{inertial}$ and  $E_{prior}$ are the photometric error of non-coplanar points (section \ref{alg: photometric error of non-coplanar point}), the photometric error of coplanar points (section \ref{alg: photometric error of coplanar point}), the inertial error term  (section \ref{alg: photometric error of inertial error}), and the prior from marginalization operator (section \ref{alg: Marginalization with coplanar constraint}), respectively. $\lambda$ is the weight between visual photometric error and inertial error.

\subsection {Coplanarity Detection}\label{alg: coplanarity detection}
The planar regularities are detected from the 3D mesh of the surrounding environment \cite{rosinol2019incremental, li2020leveraging}. The method first builds 2D Delaunay triangulation with the feature points in the image, and then projects the triangular regularities to 3D landmarks. The 3D mesh is formed by organizing landmarks into 3D patches. In the direct method VIO, as shown in Fig. \ref{fig:cover_grap} (c),  2D Delaunay triangulation is generated in the depth map.  The reason for this is that the depth map is composed of active points whose depth has converged, ensuring the accuracy of 3D mesh. Moreover, the depth map is anchored to the multi keyframes in the sliding window, avoiding the generation of 2D Delaunay triangulation frame by frame, which limits memory usage.  There might be invalid 3D triangular patches generated from depth map. We use the method in \cite{rosinol2019incremental}  to filter out these outliers, which ensures that 3D triangular patches should not be particularly sharp triangles, such as triangles with the aspect ratio high than 20 or an acute angle smaller than 5 degrees. 

The gravity direction of VIO can be used to improve the efficiency of plane detection. Thereby, we only detect the planes that are either vertical (i.e., walls, the normal is perpendicular to the gravity direction) or horizontal (i.e., floor, the normal is parallel to the gravity direction), which are commonly found in man-made environments. For horizontal plane detection, in most cases, the only horizontal plane we observed is the floor. Therefore, we only detect and optimize one horizontal plane at a time to ensure that it can be tracked for a long time. The specific strategy is:  all triangular patches that are parallel for the gravity direction are collected. Then we build a 1D histogram of the average height of triangular patches. After statistics, a Gaussian filter is used to eliminate multiple local maximums as used in \cite{rosinol2019incremental}. Finally, we extract the local maximum of the histogram and consider it to be the horizontal plane when it exceeds a certain threshold ($\sigma_t = 20$). 

For vertical plane detection, a 2D histogram is built, where one axis is the azimuth $\phi$ of the plane's normal vector, and the other axis is the distance $d$ from the origin to the plane. The histogram is divided into $n_\phi \times n_d$ bins with a bin size $\delta\phi \times \delta d$. We only count the triangular patches that the normal vector is perpendicular to the gravity direction. When the triangular patches fall into the corresponding bin, the number of  bin will be increased by 1. After statistics, we take the candidates with more than 20 inliers.

\subsection{Photometric Error of Non-coplanar Point } \label{alg: photometric error of non-coplanar point}
The direct method is based on the  photometric invariance hypothesis to minimize the photometric error. In a similar way as \cite{engel2017direct}, the photometric error for a non-coplanar point $\mathbf{p}_n$ with inverse depth $d_{p_n}$ in the host image  $I_h$ observed by the target image $I_t$ is defined as:
\begin{flalign} 
  E_{\mathbf{p}_n} = \sum\limits_{\mathbf{p}_n \in \mathcal{N}_{\mathbf{p}_n}} w_{\mathbf{p}_n} \|
  \left( I_t\left[\mathbf{p}_n'\right] - b_t\right) - \frac{t_t e^{a_t}}{t_he^{a_h}}\left(I_h\left[\mathbf{p}_n\right]-b_h\right) \|_{\gamma} \label{fla: photometric error}
 \end{flalign}
where $t_h$, $t_t$ are the exposure times of the respective image $I_h$ and $I_t$. $a_h$, $a_t$, $b_h$ and $b_t$ denote the affine illumination transform parameters. $\mathcal{N}_{\mathbf{p}_n}$ represents a small set of pixels around the point $\mathbf{p}_n$.
$w_{\mathbf{p}_n}$ stands for a gradient-dependent weighting and $\| \cdot\|_{\gamma}$ indicates the Huber norm. $\mathbf{p}_n'$ is the point projected into $I_t$, which is obtained by  (\ref{fla: point transform}). Therefore the sum of the photometric error of non-coplanar points  observed by all the keyframes in the sliding window is defined as:
 \begin{flalign} 
	E_{point} = \sum\limits_{i \in \mathcal{F}}  \sum\limits_{\mathbf{p}_n \in \mathcal{P}_i} \sum\limits_{j \in obs(\mathbf{p}_n)} E_{\mathbf{p}_n}  
\end{flalign}
where $\mathcal{F}$ is the keyframes in the sliding window, $\mathcal{P}_i$ is the set of non-coplanar points in the host keyframe $i$, and $obs(\mathbf{p}_n)$ is a set of observations of $\mathbf{p}_n$ in the other keyframes. 
\begin{figure}[t]
	\centering
	\includegraphics[width=1\linewidth]{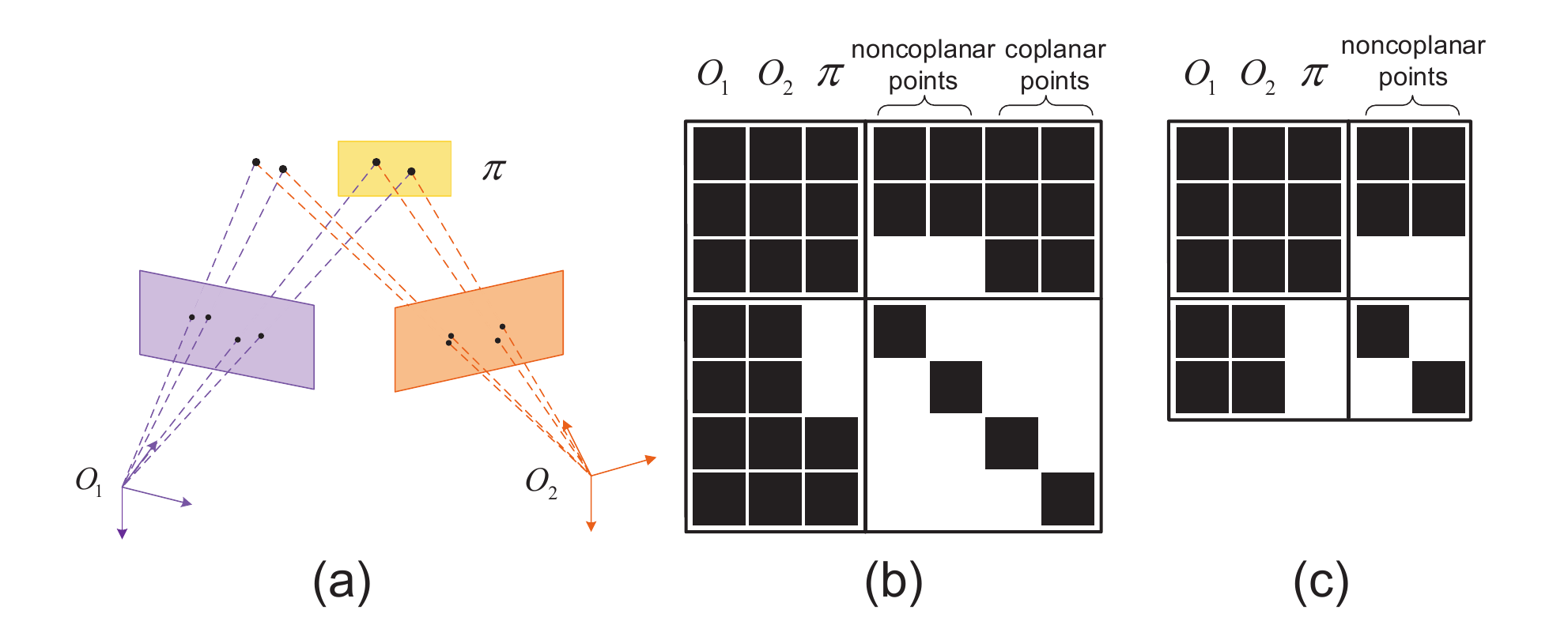}
	\caption{Hessian matrix of the coplanar point residuals and non-coplanar point residuals. 2 coplanar points and 2 non-coplanar points are observed by camera $O_1$ and camera $O_2$, and the coplanar points lie on the plane $\mathbf{\pi}$ (a); traditional residuals of coplanar points need to optimize the inverse depths, camera poses and plane parameters, which leads to the huge dimension of the hessian matrix in the optimization (b); the residuals of coplanar points in this paper only optimize the camera poses and plane parameters, which reduces the dimension of the hessian matrix (c).}
	\label{fig:hessian matrix of coplanar points}
\end{figure}
\subsection{Photometric Error of Coplanar Point } \label{alg: photometric error of coplanar point}
Assuming that the plane $\bm{\pi}_w$ detected from the 3D mesh is expressed in the world frame, the photometric error of the coplanar points can be constructed with the plane constraint equation. Suppose the 3D point corresponding to $\mathbf{p}_p$ in $I_h$ lies on the plane $\bm{\pi}_c$ transformed into the host image frame, the point $\mathbf p_p$ satisfies the coplanar equation:
\begin{flalign} \label{fla: plane_constraint}
\mathbf{n}_{\pi_c}^{\rm T} \Pi_c^{-1}(\mathbf{p}_p, 1)/d_p + d_{\pi_c} = 0
\end{flalign}   
where $\Pi_c^{-1}(\mathbf{p}_p, 1)$ represents the point on the normalized image plane.  $\bm{\pi}_c$ is the plane in the host image frame which can be obtained by (\ref{fla: transformation matrix of plane}). $\mathbf{n}_{\pi_c}$ and $d_{\pi_c}$  are the normal vector and distance of $\bm{\pi}_c$. Substituting the inverse depth $d_p$ of (\ref{fla: plane_constraint}) which is regularized by plane constraints into (\ref{fla: point transform}), the projection point $\mathbf{p}_p'$ of planar points is given by: 
\begin{flalign} 
	\begin{array}{c}
		\mathbf{p}_p' = \Pi_c\left(\mathbf{R}_{c_tc_h}\Pi_c^{-1}\left(\mathbf{p}_p, -\mathbf{n}_{\pi_c}^{\rm T} \Pi_c^{-1}(\mathbf{p}_p, 1)/d_{\pi_c}\right) + \mathbf{t}_{c_tc_h}\right)
	\end{array}\label{fla: coplanar_point transform}
\end{flalign} 
The photometric error of coplanar points is obtained by substituting $\mathbf{p}_p'$ into (\ref{fla: photometric error}), which is the same as that of non-coplanar points. The sum of the photometric error of coplanar points observed by all the keyframes in the sliding window can be written as:
 \begin{flalign} 
	E_{point}' = \sum\limits_{i \in \mathcal{F}}  \sum\limits_{\mathbf{p}_p \in \mathcal{C}_i} \sum\limits_{j \in obs(\mathbf{p}_p)} E_{\mathbf{p}_p}  
\end{flalign}
where $\mathcal{C}_i$ is the set of coplanar points in the host keyframe $i$, $obs(\mathbf{p}_p)$ is a set of observations of the $\mathbf{p}_p$ in the other keyframes.

For the optimization method, deriving the analytical Jacobian of estimated variables with the chain derivation method can significantly improve the efficiency of the optimizer. For a single photometric error, the Jacobian corresponding to the state variables can be decomposed as \cite{engel2017direct}:
\begin{flalign}
\mathbf{J}_{E_{\mathbf{p}_p}} = [\mathbf{J}_I \cdot \mathbf{J}_{geo}, \mathbf{J}_{photo}]
\end{flalign}
where $\mathbf{J}_I$ represents the image gradient. $\mathbf{J}_{geo}$ denotes the Jacobian of geometric parameters ($\textbf{T}_{wi_h}$, $\textbf{T}_{wi_t}$, $\bm{\pi}_w$), and $\mathbf{J}_{photo}$ indicates the Jacobian of photometric parameters ($a_h$, $a_t$, $b_h$, $b_t$). Please refer to the appendix for the specific forms of $\mathbf{J}_{E_{\mathbf{p}_p}}$. It is worth noting that the coplanar constraints used in this paper do not optimize the depth of optimization points simultaneously. As shown in Fig.\ref{fig:hessian matrix of coplanar points}, compared with the traditional point-to-plane constraints \cite{rosinol2019incremental, li2020leveraging}, the efficiency of the algorithm can be improved significantly by reducing the dimension of the Hessian matrix in the optimizer. 

\subsection{Inertial Error}  \label{alg: photometric error of inertial error}
To construct the inertial error with angular velocity and linear acceleration measurements, the pre-integration method proposed in \cite{forster2015imu} is used to handle the high frequency of IMU measurements. 
This gives a pseudo-measured value between consecutive keyframes.
Given the previous IMU state $\mathbf{s}_i^I:=\{\mathbf{t}_{wi}, \mathbf{R}_{wi}, \mathbf{v}_{wi}, \mathbf{b}_{ai}, \mathbf{b}_{gi}\}$, which contains translation, rotation, velocity in the IMU frame, and gyroscope and accelerometer biases, the preintegration measurements provide us with the prediction state $\widehat{\mathbf{s}}_j^I$ for the following state $\mathbf{s}^I_j$ as well as a covariance matrix $\widehat{\mathbf{\Sigma}}_j$. The resulting inertial error function penalizes deviations from the current estimated state to the predicted state.
 \begin{flalign} 
  E_{inertial} = \left(\widehat{\mathbf{s}}_j^I \boxminus \mathbf{s}_j^I\right)^T \widehat{\mathbf{\Sigma}}_j^{-1}\left(\widehat{\mathbf{s}}_j^I \boxminus \mathbf{s}_j^I\right)
 \end{flalign}
The subtraction operation $\mathbf{s}_i \boxminus \mathbf{s}_j$ is defined as $\log\left(\mathbf{R}_i\mathbf{R}_j^{-1}\right)$ for rotation and a regular subtraction for vector values.

\subsection{Marginalization about Coplanar Constraints}\label{alg: Marginalization with coplanar constraint}
To balance the efficiency and accuracy, the marginalization method is used in VO/VIO \cite{usenko2019visual, engel2017direct, von2018direct, von2022dm}. When a new image frame is added to the sliding window, all the variables corresponding to the older frame are marginalized using the Schur complement \cite{usenko2019visual}. The marginal keyframe is selected similarly to the criteria in \cite{engel2017direct}, which considers the luminance change of the images and the geometry distribution of the poses. Meanwhile, to maintain the consistency of the system, once the variable is connected to the marginalization factor, the First Jacobian Estimation (FEJ) \cite{huang2009first} is used to fix the variable's linearization point at the same value.

In the sliding window optimization, maintaining too many historical planes in the optimization will seriously affect the efficiency of the optimizer. The historical plane should also be "marginalized" to form the plane's prior factor.  For simplicity, the plane prior factor generated by marginalization is replaced by a plane-distance cost, which can be expressed by the distance from the prior plane $\bm{\pi}'= (\phi',  \psi', d')$ to the currently optimized plane $\bm{\pi} = ( \phi ,  \psi , d)$:
\begin{flalign} \label{fla: plane prior}
	 E_{\mathbf{\pi}_p} = w_n  \|\left[ \phi', \psi', d' \right]^{\rm T}- \left[ \phi ,  \psi , d \right]^{\rm T}\|_{\sum_{\mathbf{\pi}}}^2
\end{flalign}
where $\sum_{\mathbf{\pi}}$ is the covariance matrix. $w_n$ denotes the number of the coplanar constraints corresponding to the plane. If the plane is marginalized, a prior plane $\bm{\pi}'$ is formed. When the plane is observed again, it is reactivated and optimized with prior constraint (\ref{fla: plane prior}) in the sliding window.

\section{Experiments} 
\label{sec:experiment}
In this section, simulation experiments are first used to verify the effectiveness of our method, and then the evaluation on the real EuRoC MAV dataset \cite{Burri2016} and TUM VI dataset \cite{Schubert2018} demonstrates the accuracy and efficiency. We provide a video \footnote{https://youtu.be/h-3fP6VP0\_k} to reflect the results of the experiments more intuitively. The related code about our analytical Jacobian of coplanar parametric representation is published on GitHub to facilitate communication\footnote{
https://github.com/boxuLibrary/PVI-DSO-SIM}.  
We run the system in the environment with Intel Core i7-9750H@ 2.6GHz, 32GB memory.
\subsection{Quantitative Evaluation}
To evaluate the performance of the coplanar constraints in VIO system, we re-implemented a VIO based on DSO \cite{engel2017direct}, which is denoted as VI-DSO-RE to distinguish it from VI-DSO \cite{von2018direct} without open source code, and the coplanar constraints are added into VI-DSO-RE, which is denoted as PVI-DSO.  To verify the effectiveness of coplanar constraints in the direct method, we evaluate the performance of VI-DSO, VI-DSO-RE and PVI-DSO, and also compare PVI-DSO with the state-of-the-art VIOs fusing coplanar constraints: PVIO  \cite{li2019robust} with multi-plane priors and mesh-VIO \cite{rosinol2019incremental}, which fuses the planar regularities generated by 3D meshes.

The simulations and real experiments are conducted, in which we perform  $\mathbf{SE}(3)$ alignment against the groundtruth to get the Root Mean Square Error (RMSE) of the translation and rotation error. The scale error is computed with $| 1-s|$, where $s$ is obtained from $Sim(3)$ alignment. For the TUM VI dataset, since the trajectory lengths can vary greatly, the drifts in \% are also computed with $ \rm \frac{RMSE \cdot 100}{length}$ as in \cite{von2022dm}.  Unless otherwise states all methods are evaluated 10 times for the EuRoC MAV dataset and TUM VI dataset on each sequence and the medium results are reported.

\begin{figure}[hptb]
	\centering
	\includegraphics[scale=0.15]{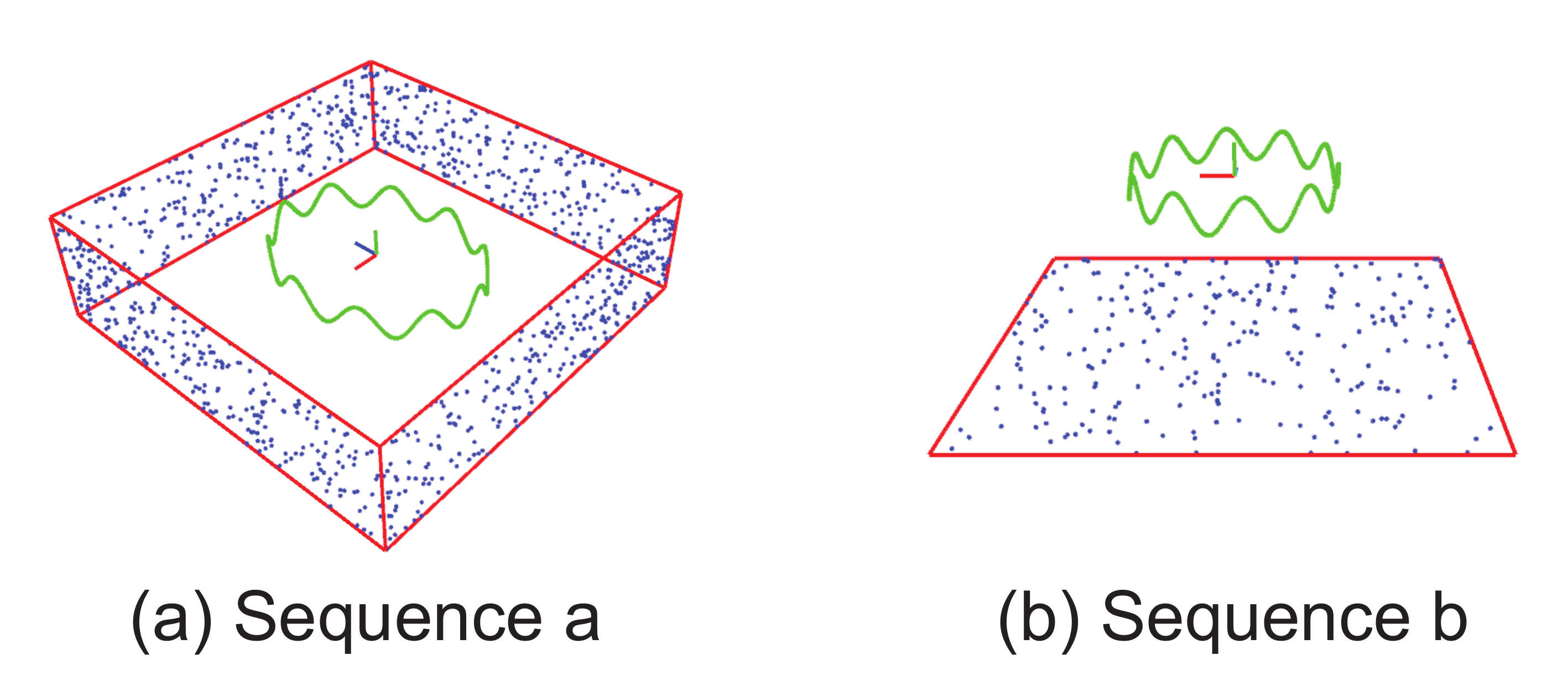}
	\caption{
	Two simulation environments are illustrated, where points in blue are distributed on the vertical planes (a) and the horizontal plane (b). The camera follows green trajectories. }
	\label{fig:synthetic sequences}
\end{figure}
\subsubsection{Simulation Experiment}
Two simulation sequences with ideal coplanar environments are created to evaluate the efficiency concerning performance under different parametric representations. As shown in Fig.\ref{fig:synthetic sequences}, we simulated camera motion with 10 Hz, forming an ellipse trajectory with sinusoidal vertical motion. The long-semi axis and short axis of the ellipse trajectory are 4 m and 3 m, respectively. The number of landmarks in each plane is limited to 250, which consists of the vertical plane sequence and horizontal plane sequence. The Gaussian noise is added to the landmark observations ($\sigma_p = 1$ pixel), plane parameters ($\sigma_{n} = 5$ degrees, $\sigma_{d}$ = 0.3 m), and camera poses ($\sigma_R = 5$ degrees, $\sigma_{t}$ = 0.3 m) to simulate the real  observation environment. The comparison method includes: Point, PP (-L) and PP (-T), where Point denotes the point-based method, the implementation of which is obtained from open-sourced SLAM VINS \cite{VINS}. Both PP (-L) and PP (-T) use coplanar constraints in the optimization module, but in different ways.  PP(-L) fuses more constraints between point-to-plane with the loosely coupled method as in \cite{rosinol2019incremental, li2020leveraging}.  Whereas PP (-T) adopts the parametric representation used in this paper to fuse plane constraints in a tightly coupled method. Considering it is difficult to simulate photometric error,  Point, PP (-L), and PP (-T) all adopt the geometric reprojection error instead to verify the performance of different parametric representations in the simulation experiment.

\begin{figure}[t!]
	\centering
	\includegraphics[scale=0.55]{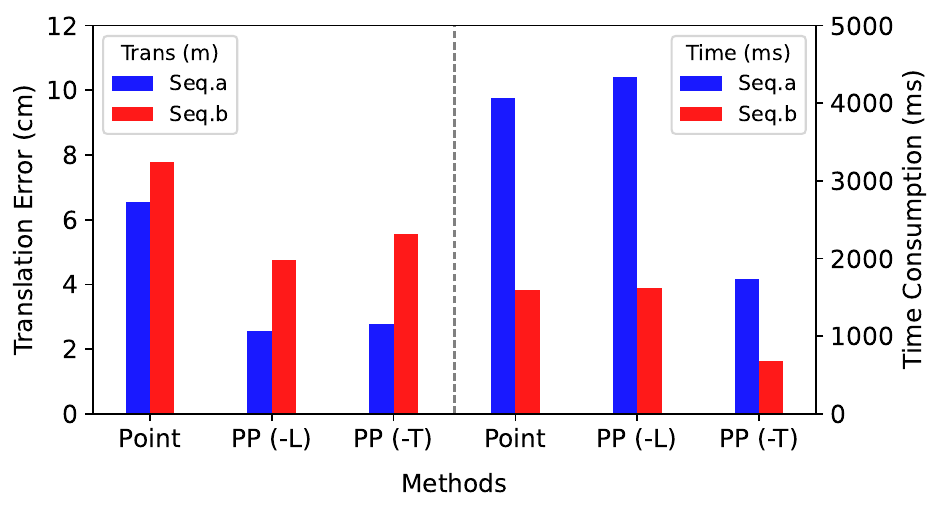}
	\caption{Translation error (m) and running time (ms) of Point, PP(-L) and PP(-T) methods on two simulation sequences.}
	\label{fig:synthetic results}
\end{figure}

As shown in Fig.\ref{fig:synthetic results}, compared with the Point method, the translation error of PP (-L) and PP (-T) with coplanar constraints decreases on both sequences, which demonstrates that the performance of the VIO can be improved with the structural information in the environment. The accuracy difference between PP (-L) and PP (-T) is marginal, whereas the effectiveness of PP (-T) is improved significantly, which is about 2.3 times and 2.4 times faster than Point and PP (-L), respectively. This is because PP (-T) reduces the dimension of the Hessian matrix in the optimization without estimating the state variables of coplanar points.

\begin{table*}[htbp]
	\label{tab:1}
    \renewcommand\arraystretch{0.8}
	\begin{center}
	\caption{Trajectory error (m) of different methods on EuRoC dataset. In \textbf{bold} the best results. In \textcolor{blue}{blue} the second best results}
		\label{tab:ATE Trans of EuRoC}
		\centering
		\setlength{\tabcolsep}{2.0mm}{
        \begin{threeparttable}
			\begin{tabular}{cc|ccccccccccc}
				\toprule
				\multicolumn{2}{c|}{Sequence} &MH1&MH2&MH3&MH4&MH5&V11&V12&V13&V21&V22&V23  \\\midrule
                 \multirow{3}{*}{\makecell[c]{VI-DSO\tnote{1} \\ } } 
                &RMSE& \bf 0.062  & \bf 0.044& 0.117  & \textcolor{blue}{0.132} & \bf 0.121 & 0.059 & \bf 0.067 & 0.096 & \bf 0.040 & 0.062 & 0.174 \\
				&RMSE gt-scaled & 0.041 & 0.041  & 0.116 & 0.129 & 0.106 & 0.057 &0.066 & 0.095 & 0.031 & 0.060 & 0.173 \\
				&Scale Error(\%) &\textcolor{blue}{1.1}  & 0.5 & 0.4 &  \textcolor{blue}{0.2} & 0.8 & \bf 1.1 & \bf 1.1 &  \bf 0.8 & 1.2 &  \bf 0.3 & \textcolor{blue}{0.4} \\\midrule
    
				\multirow{3}{*}{\makecell[c]{VI-DSO-RE \\ } } 
				&RMSE & 0.081 &  \textcolor{blue}{0.046}  & \bf 0.054 & 0.137 & 0.205 & \bf 0.051 &0.108  & \textcolor{blue}{0.095} & 0.060& \textcolor{blue}{0.057} & \textcolor{blue}{0.138} \\
                &RMSE gt-scaled & 0.059 &0.046  & 0.053 & 0.137 & 0.202 & 0.041 &0.103  & 0.091 & 0.060 & 0.046 & 0.138 \\
				&Scale Error(\%) &1.3  & \bf 0.1 & \bf 0.1 & \bf 0.1 & \textcolor{blue}{0.5} & 1.7 &  \textcolor{blue}{1.9} &  1.5 & \textcolor{blue}{0.2} & 1.8 & \bf 0.1 \\\midrule

				\multirow{3}{*}{PVI-DSO}
                & RMSE &  \textcolor{blue}{0.073} &  \textcolor{blue}{0.046} & \textcolor{blue}{0.055} & \bf 0.114 & \textcolor{blue}{0.175} & \textcolor{blue}{0.056} & 0.100 & \bf 0.087 & \textcolor{blue}{0.050} & \bf 0.044 & \bf 0.111 \\
				& RMSE gt-scaled & 0.054 & 0.042 &0.054 & 0.107 & 0.175 & 0.052 & 0.094 & 0.085 & 0.050 & 0.044 & 0.109 \\
				& Scale Error(\%)  & \bf 0.7 &  \textcolor{blue}{0.4} & \textcolor{blue}{0.3} & 0.5 &\bf 0.2 &  \textcolor{blue}{1.2} & \textcolor{blue}{1.9} & \textcolor{blue}{0.9} &  \bf 0.1 & \textcolor{blue}{1.7} &  1.0 \\ \midrule
    
                %\makecell[c]{ DM-VIO}& RMSE  & 0.065 & \textcolor{blue}{0.044} & 0.097 & 0.102 & \bf 0.096 & \bf 0.048 &\bf 0.045  & \bf 0.069 & \bf 0.029 & 0.050 & \textcolor{blue}{0.114} \\ \midrule
                
                \makecell[c]{ mesh-VIO\tnote{2}}& RMSE  & 0.145 & 0.130 & 0.212 & 0.217 & 0.226 & 0.057 & \textcolor{blue}{0.074}  & 0.167 & 0.081 & 0.103 & 0.272 \\ \midrule

                \makecell[c]{ PVIO\tnote{3}}& RMSE  & 0.163 & 0.111 & 0.119 & 0.353 & 0.225 & 0.082 &0.113  & 0.201 & 0.063 & 0.157 & 0.280 \\ \midrule
			\end{tabular}
        \begin{tablenotes}    
        \footnotesize 
        \item[1] results taken from \cite{von2018direct}.
        \item[2] results taken from \cite{rosinol2019incremental}.     
        \item[3] results taken from \cite{li2019robust}.   
         
        \end{tablenotes}           
        \end{threeparttable}     
		}
	\end{center}
\end{table*}

\subsubsection{EuRoC Dataset}
\begin{figure}[htpb]
	\centering
	\includegraphics[scale=0.25]{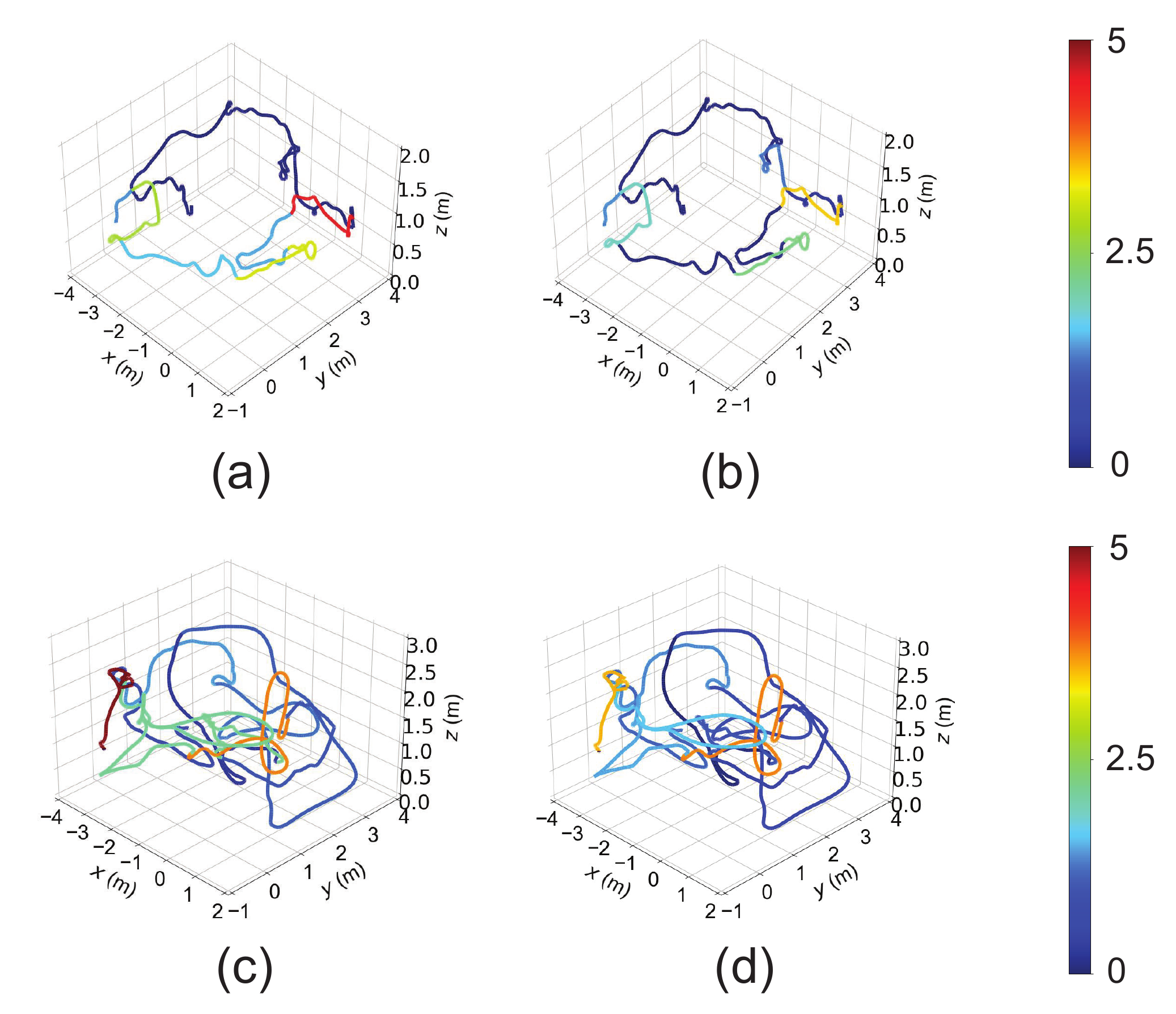}
	\caption{
    Scale error comparison of VI-DSO-RE  and PVI-DSO. The trajectories are divided into multiple segments, and aligned with groundtruth to calculate scale error, respectively. The two colorful trajectories of the left column are running with VI-DSO-RE on the (a) V21 and (c) V22 datasets. The right two trajectories are the results of PVI-DSO on the (b) V21 and (d) V22 datasets. Colors encode the corresponding scale errors (\%).}
	\label{fig:map of different scale}
\end{figure}

The EuRoC micro aerial vehicle (MAV) dataset consists of two scenes, the machine hall and the ordinary room, 
which contain different motion and visual scenes. We compared the positioning accuracy of VI-DSO, VI-DSO-RE, PVI-DSO, PVIO and mesh-VIO. As shown in Tab. \ref{tab:ATE Trans of EuRoC}, the average RMSE of VI-DSO, VI-DSO-RE and PVI-DSO are 0.089m, 0.094m and 0.083m, respectively. VI-DSO adopts dynamic marginalization to maintain the consistency of the marginalization prior. VI-DSO-RE does not use any marginalization tricks, so the accuracy of VI-DSO is slightly higher than that of VI-DSO-RE. Furthermore, the accuracy of PVI-DSO leveraging planar regularities outperforms VI-DSO and VI-DSO-RE, which demonstrates that the prior structural information fused into VIO based on direct method suppresses the rapid divergence of the VIO, thereby reducing the cumulative error generated by long-time operation of VIO. By comparing the scale error of VI-DSO-RE and PVI-DSO, the average scale error of PVI-DSO decreases by 4\%. For further analysis, we divided the trajectories of V21 and V22 into multiple segments with 10s intervals to compute the scale error separately. As shown in Fig. \ref{fig:map of different scale}, the scale error of PVI-DSO decreased in some segmented trajectories compared to VI-DSO-RE, which indicates the consistency of the global scale estimation is improved after introducing the structural information. Finally, from the experimental results of PVI-DSO and two methods that introduce the planar information into VIO based on the indirect method, we can observe that our method is significantly better than PVIO and mesh-VIO. This is because VIO based on the direct method utilizes more visual information in the scene to estimate the state variables. Moreover, the denser map constructed by PVI-DSO contains rich structural information, which promotes the detection of the planar regularities in the map, so as to improve the positioning accuracy better after introducing the planar constraints. 
\begin{figure}[b]
	\centering
	\includegraphics[scale=0.23]{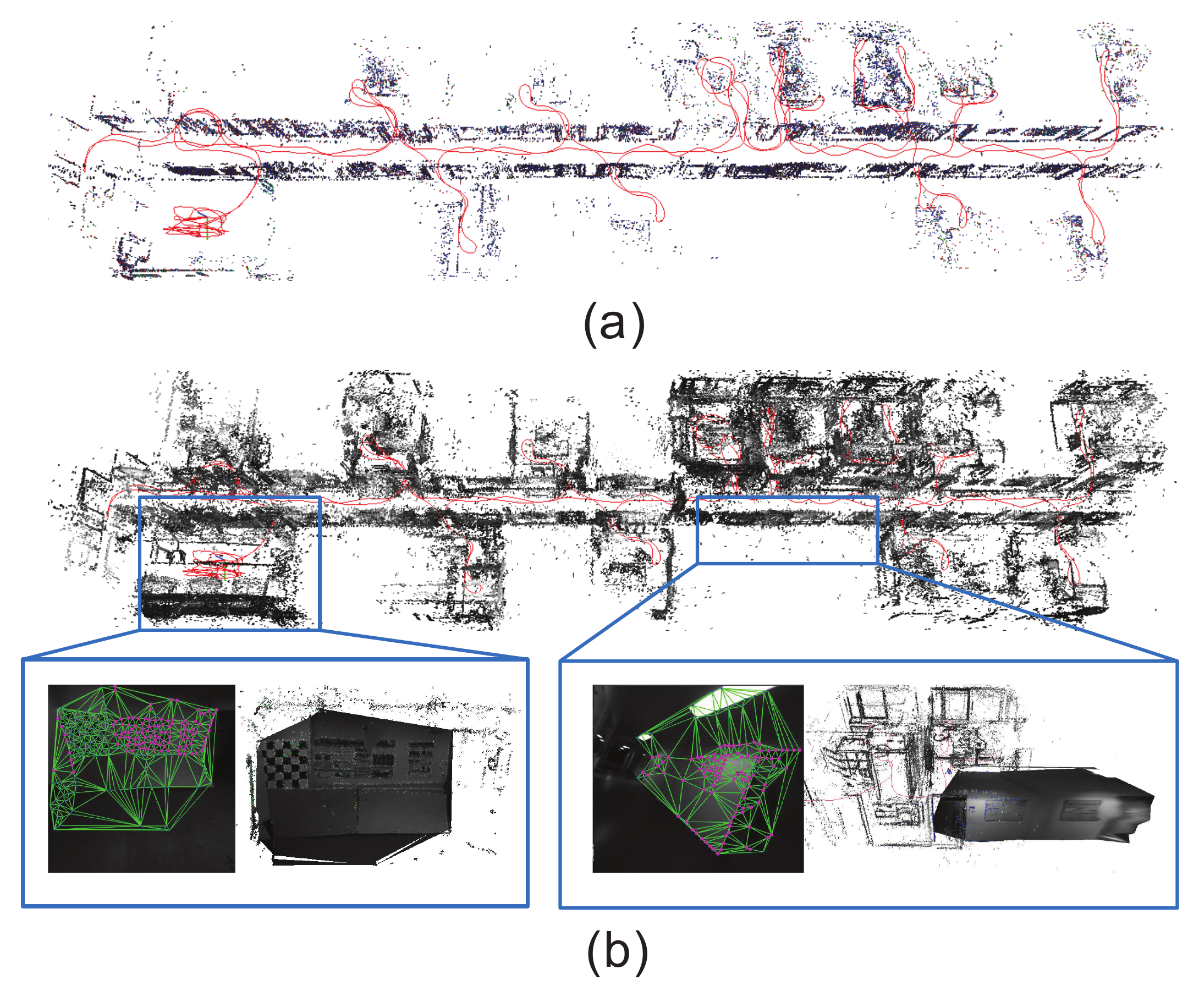}
	\caption{
	Reconstructed map and trajectory of corridor3 sequence with PVI-DSO. (a) is the 3D coplanar points  on the vertical planes and the horizontal planes of the corridor. (b) is the reconstructed denser map. Two sub-images show the 2D Delaunay triangulation and the corresponding 3D mesh in the scene, where the pink points in the image are the detected 2D coplanar points. }
	\label{fig:map of tum vi}
\end{figure}

\begin{table}[htpb]
    \renewcommand\arraystretch{1.0}
	\label{tab:1}
	\begin{center}
	\caption{RMSE comparison of different methods on TUM VI dataset.  Translation (m) and rotation (rad) errors are listed as follows. In \textbf{bold} the best results}
	\label{tab:ATE Trans and Rots on TUM VI}
	\centering
	\setlength{\tabcolsep}{1.0mm}{
\begin{tabular}{c|cccccc|c}
\toprule
\multirow{2}{*}{sequence} & \multicolumn{2}{c}{PVIO} & \multicolumn{2}{c}{VI-DSO-RE} & \multicolumn{2}{c|}{PVI-DSO } & \multirow{2}{*}{length}\\
 \cmidrule(r){2-3}   \cmidrule(r){4-5}  \cmidrule(r){6-7}
      & trans        & rot       & trans         & rot        & trans         & rot&        \\ \midrule
corridor1   &  0.271 &      0.379      &    0.247 &  \bf 0.094     &  \bf 0.104   &   0.135         &  305      \\
corridor2  &  \bf 0.442  &  0.889         &   0.526 & \bf 0.414     &     0.457          &  0.432           & 322       \\
corridor3  &  4.958 & 3.041       &      \bf 0.167         & 0.169           &  0.199   &    \bf 0.154         & 300       \\
corridor4  &  0.760 & 0.574          & \bf 0.104              &    0.126        & 0.119              &  \bf 0.105           & 114       \\
corridor5  &  0.718  & 1.021          &  0.439             &  0.428          & \bf 0.119        &    \bf 0.025         & 270    \\
magistrale1&  5.061 &  0.450         &  \bf 1.397&    \bf 0.174   & 1.487              &   0.200          & 918  \\
magistrale2& - & - & 
\bf 1.002              & \bf 0.064           &  1.334             &   0.122          & 561  \\
magistrale3& - & - & 1.150              &  0.095  &  \bf 0.847    &     \bf 0.090        & 566  \\
magistrale4&  2.850 &  0.218         &  2.073   & 0.266      & \bf 1.788              &   \bf 0.194        & 688  \\
magistrale5&  2.998 &   0.532        &  0.284   &  \bf 0.021          & \bf 0.202              &   0.025          & 458  \\
magistrale6&   6.068 & 2.465           &  2.470 &  0.654 &  \bf 2.268  &   \bf 0.600          & 771  \\
\textbf{avg drift} \%&   0.580 & 0.312           & 0.161 & 0.060  & \bf 0.140               & \bf 0.046            & normalized  \\
\midrule
\end{tabular}
		}
	\end{center}
\end{table}

\subsubsection{TUM VI Dataset} 
We also evaluate our proposed system on the two sequences (corridor and magistrale) of the TUM VI dataset. The sequences contain a large number of images with motion blur and photometric changes, which is challenging for the direct method. In Fig. \ref{fig:map of tum vi}, we reveal the reconstructed map and the extracted coplanar points on the ground and walls running PVI-DSO. The drift of VIO can be effectively suppressed using the structural information in the scene. From Tab. \ref{tab:ATE Trans and Rots on TUM VI}, it can be seen that compared with PVIO, no matter VI-DSO-RE or PVI-DSO, the translation error and rotation error decrease significantly. Furthermore, introducing the planar constraints into VI-DSO-RE, the average translation drift and rotation drift of PVI-DSO decrease by 13\% and 23\%, respectively, which verifies the effectiveness of our proposed method. It should be noted that there are sequences where the accuracy of PVI-DSO decreases after introducing the planar information (eg., corridor3-4, magistrale1-2). This is because the estimation of the plane parameters gradually converges during system operation. The visual observation noise may makes the plane fail to converge to accurate position,  and the inaccurate prior information can damage the performance of the system.

\subsection{ Weight Determination of Photometric Error}

\begin{figure}[htpb]
	\centering
	\includegraphics[scale=0.18]{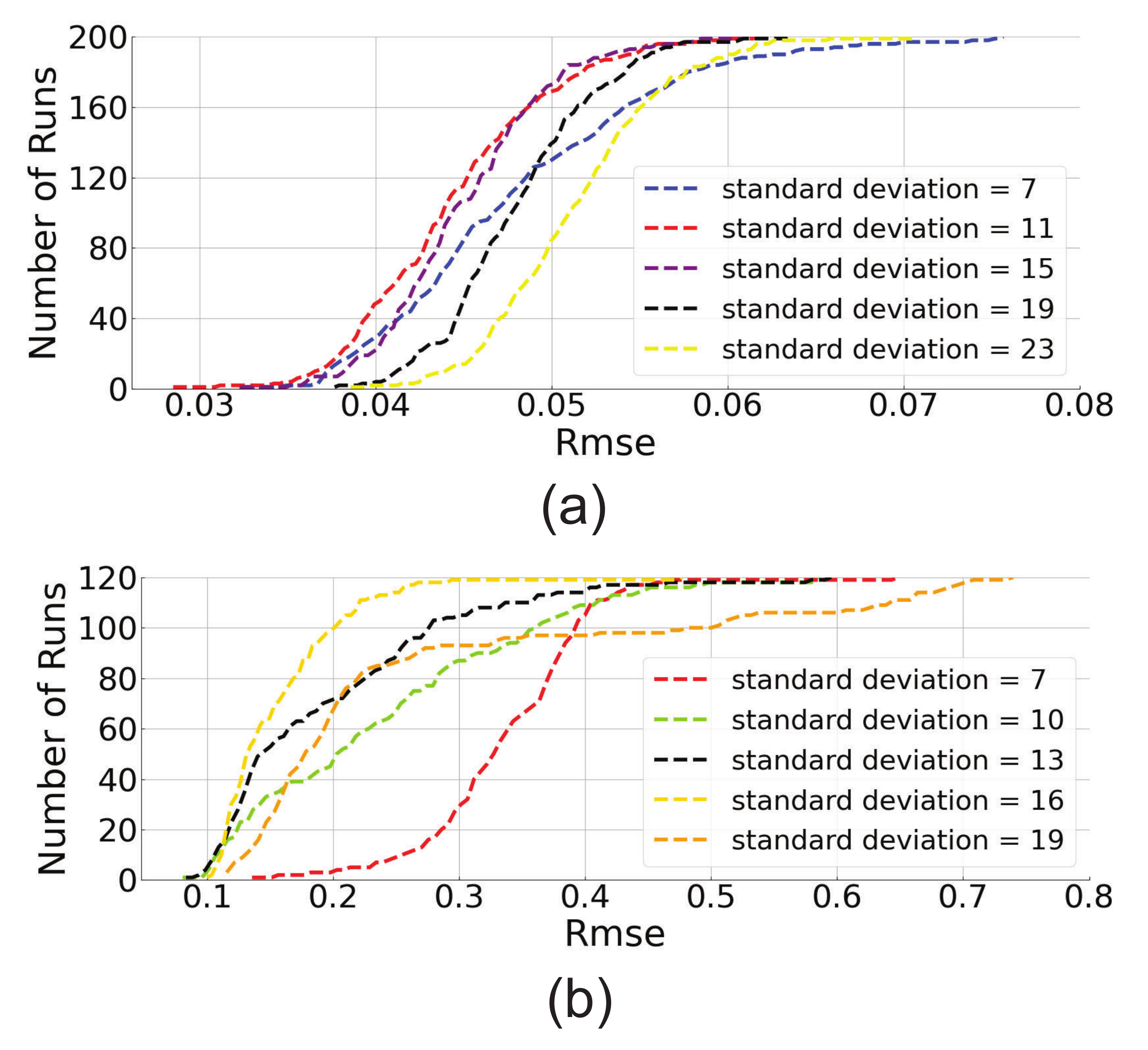}
	\caption{Cumulative error plot on the V22 sequence of EuRoC dataset (a) and corridor4 sequence of TUM VI dataset (b). The different standard deviation of photometric error affects the accuracy and robustness of pose estimation.}
	\label{fig:v2_02_standard_derivation}
\end{figure}

Since the camera's photometric calibration and distortion correction have significant influences on the standard deviations of the photometric error, it is difficult to determine the weight ratio of the photometric residuals and inertial residuals in the optimization. In this paper, the parametric study method is used to obtain the optimal standard deviation of the photometric error. After setting different standard deviations, the cumulative error curves of 200 and 120 runs on the EuRoC dataset and TUM VI dataset are  counted. According to Fig. \ref{fig:v2_02_standard_derivation}, it can be seen that the different weight ratio of photometric and inertial residual has a great impact on the positioning results of the system, and the experimental results on the TUM VI dataset are more sensitive to the different settings.  At the same time, we can also observe that in the experiments, the best performance is obtained by setting the photometric error standard deviation 11 and 16 on the EuRoC and TUM VI dataset, respectively.  

\subsection{Runtime Evaluation}
To show the time-consuming details of the algorithm, the time consuming of different modules in the algorithm is carefully counted. We calculated the time consumption of VO based on the direct method, VI-DSO-RE introducing the inertial constraints in the VO, and PVI-DSO leveraging the planar regularities in the VI-DSO-RE. Considering the real-time performance of the system, we set the number of extracted pixels in the keyframe to 800, the size of the sliding window to 7 and the maximum number of iterations to 5.

The computation cost (in milliseconds) of different modules for running on EuRoC's V11 sequence is shown in Tab. \ref{tab:running time}. In pure visual VO, about two-thirds of the time is used for optimization, and the rest time is spent on the visual feature processing module in direct VO, including pixels tracking, candidates selection, points activation, pixels extraction, etc. After introducing the IMU measurements into VO, the total time consumption of VI-DSO-RE increases slightly. For PVI-DSO, plane detection and mesh construction have little burden on the system operation, which only cost 0.72 ms and 1.08 ms, respectively. Whereas compared with VI-DSO-RE, the time cost of the optimization and marginalization is decreased. The reason is that PVI-DSO does not need to optimize the coplanar points, which reduces the dimensions of the matrix to be solved. The statistics shows that in the sliding window, the average number of the state variables (plane number $\approx$ 2, point number $\approx$ 1182) of PVI-DSO is fewer than that (point number $\approx$ 1422) of  VI-DSO-RE.

\begin{table}[htbp]
	\label{tab:3}
	\begin{center}
		\caption{Mean execution time (Unit: millisecond) of  VO, VI-DSO-RE, and PVI-DSO running on the sequence V11.}
		\label{tab:running time}
		\centering
		\setlength{\tabcolsep}{1.6mm}{
			\renewcommand{\arraystretch}{1.2}
			\begin{threeparttable}
			\begin{tabular}{cccc}
			\hline
			\multicolumn{1}{c}{Module} & \multicolumn{1}{c}{VO} & \multicolumn{1}{c}{VI-DSO-RE} & \multicolumn{1}{c}{PVI-DSO} \\ \hline
			\texttt{Plane detection } &      0                  &    0                   &  0.72                     \\
			\texttt{mesh Creation}    &       0                 &    0                    &  1.08                   \\
            \texttt{Optimization}      &   18.83            &    19.40                      &    17.49        \\    
			% \texttt{ Cost Function Construction\tnote{1}}      &   16.44            &    16.99                      &    17.74                   \\
			% \texttt{\makecell[c]{Problem Solving\&Var Updating\tnote{1}}}      &    2.39            &    2.41                      &    2.16                   \\
			\texttt{Marginalization}&     0.77               &     0.79                  &   0.72                    \\ \hline
			\texttt{Total}                        &   31.97                &    32.54                &   32.36                    \\ \hline
		\end{tabular}
		% \begin{tablenotes}    %这行要添加， 从这开始
        %\footnotesize               %这行要添加
        % \item[1] The optimization contains cost function construction, problem solving, and state variables updating.          
        % \end{tablenotes}            %这行要添加
         \end{threeparttable}       %这行要添加，到这里结束
		}
	\end{center}
\end{table}

\section{Conclusion}
In this paper, we present a direct sparse visual-inertial odometry that leverages planar regularities, which is called PVI-DSO. The denser map reconstructed from the VIO based on the direct method provides rich structure information about the scene, which makes it easy to extract plane regularities from the 3D meshes built on the point cloud. The introduction of  geometric information regulars the 3D map, which simultaneously benefits the pose estimation.  For the efficiency of the optimization, a tightly coupled coplanar constraint expression is used and the analytical Jacobian of the linearization form is derived. Extensive experiments demonstrate that our system outperforms the state-of-the-art visual-inertial odometry in pose estimation. In the future, the engineering drawing will be introduced in the VIO based on the direct method to provide the accurate prior information, which will be used to further improve the accuracy and robustness of the positioning.

\section*{Acknowledgment}
We would like to thank Dr. Yijia He for the helpful discussion. This work was supported by the National Natural Science Foundation of China (Major Program, Grant No. 42192533) and the National Key Research and Development Program of China (Grant No. 2021YFB2501100).

\appendices
\section*{APPENDIX}
This section introduces the specific expression of the Jacobian in the coplanar constraint equation. The linearized photometric observation equation for a single coplanar point $\mathbf{p}$ can be written as:
\begin{equation}
\begin{aligned}
r_k&=\left( I_t\left[\mathbf{p}'\right] - b_t\right) - \frac{t_t e^{a_t}}{t_he^{a_h}}\left(I_h\left[\mathbf{p}\right]-b_h\right)  \\
&=\mathbf{J}_{photo} \delta \mathbf{x}_{photo} + \mathbf{J}_I \cdot \mathbf{J}_{pose} \delta \mathbf{x}_{pose} + \mathbf{J}_I \cdot \mathbf{J}_{plane}\delta \mathbf{x}_{plane}
\end{aligned}
\end{equation}
where $\delta \mathbf{x}_{photo} =\left[  \begin{matrix}\delta a_h & \delta a_t & \delta b_h & \delta b_t \end{matrix}\right]^{\rm T}$ denotes the error state vector of the photometric parameters; $\delta \mathbf{x}_{pose} = \left[  \begin{matrix}\delta \mathbf{T}_{wi_h} & \delta \mathbf{T}_{wi_t} \end{matrix}\right]^{\rm T}$ represents the error state vector of the host IMU pose $\textbf{T}_{wi_h}$ and target IMU frame $\textbf{T}_{wi_t}$; $\delta \mathbf{x}_{plane} = \left[  \begin{matrix}\delta \mathbf{n}_{w} & \delta d_w \end{matrix}\right]^{\rm T}$ indicates the error state vector of plane parameters; $\mathbf{J}_I = \left[dx \quad dy\right]$ means the image gradient in $x$ and $y$ direction.  $\mathbf{J}_{photo}$, $\mathbf{J}_{pose}$ and $\mathbf{J}_{plane}$ represent the corresponding Jacobians of photometric parameters, IMU poses, and plane parameters, respectively. $\mathbf{J}_{pose}$ and $\mathbf{J}_{plane}$ constitute $\mathbf{J}_{geo}$.

The Jacobian of $\mathbf{J}_{photo}$ is same as \cite{engel2017direct}, which is given by:

\begin{equation}
\begin{aligned}
\frac{\partial r_k}{\partial \delta \left[\begin{matrix}  a_h \ a_t \ b_h \ b_t \end{matrix}\right]} = \left[\begin{matrix} 
 \frac{t_t e^{a_t}}{t_he^{a_h}}\left(I_h\left[\mathbf{p}\right]-b_h\right) \\ -\frac{t_t e^{a_t}}{t_he^{a_h}}\left(I_h\left[\mathbf{p}\right]-b_h\right) \\ \frac{t_t e^{a_t}}{t_he^{a_h}} \\ -1\end{matrix}\right]^{\rm T}
\end{aligned}
\end{equation}

The Jacobian of the host IMU frame $\textbf{T}_{wi_h}$ is written as:
\begin{equation}
\begin{aligned}
\frac{\partial r_k}{\partial \delta \mathbf{T}_{wi_h}}= \mathbf{J}_I \cdot \frac{\partial \mathbf{p}'}{ \partial \Pi_c^{-1}(\mathbf{p}', d_p)} \cdot \frac{\partial \Pi_c^{-1}(\mathbf{p}', d_p)}{ \partial \delta \bm{\xi}_{wi_h}}
\end{aligned}
\end{equation}
where $\Pi_c^{-1}(\mathbf{p}', d_p) = \left[\begin{matrix} p'_x & p'_y & p'_z \end{matrix}\right]$ represents the 3D landmarks in the target image frame. $\delta \bm{\xi}_{wi_h}$ is the minimal parametric representation of host IMU pose. 

\begin{equation}
\begin{aligned}
\frac{\partial \mathbf{p}'}{ \partial \Pi_c^{-1}(\mathbf{p}', d_p)} =
\left[\begin{matrix} 
\frac{f_x}{p'_z} & 0 & -\frac{f_x p'_x}{p'_z{^2}} \\
0& \frac{f_y}{p'_z} & -\frac{f_y p'_y}{p'_z{^2}}
\end{matrix}\right]
\end{aligned}
\end{equation}
where $f_x$ and $f_y$ are the focal length of the camera. 
\begin{equation}
\begin{aligned}
\frac{\partial \Pi_c^{-1}(\mathbf{p}', d_p)}{ \partial \delta \bm{\xi}_{wi_h}}=
\left[\begin{matrix} 
\mathbf{J}_{rot} & \mathbf{J}_{trans}
\end{matrix}\right]
\end{aligned}
\end{equation}
where $\mathbf{J}_{rot}\in \mathbb{R}^{3\times3}$ is the Jacobian of the rotation part of the host IMU frame, and $\mathbf{J}_{trans}\in \mathbb{R}^{3\times3}$ is the Jacobian of the translation part of the host IMU frame. They are given by: 
\begin{equation}
\begin{aligned}
&\mathbf{J}_{rot} = -\frac{-\mathbf{R}_{wc_t}^{\rm T} \mathbf{R}_{wi_h} \lfloor\mathbf{f}_i d_c\rfloor_{\times} t_{cp} - d_c \mathbf{f}_t \mathbf{f}_i^{\rm T} \lfloor\mathbf{R}_{wi_h}^{\rm T} \mathbf{n}_w \rfloor_{\times}  }{t_{cp} \times t_{cp}}  \\
&\mathbf{J}_{trans} = \frac{-\mathbf{f}_t \cdot \mathbf{n}_w}{t_{cp}} + \mathbf{R}_{wc_t}^{\rm T} 
\end{aligned}
\end{equation}
where $\mathbf{f}_c$ is the point observation in the normalized plane of the host image frame. $\left(\mathbf{R}_{wc_*}, \mathbf{t}_{wc_*}\right),  * \in \{ h, t \}$ indicates the rotation and translation of the host image frame and target image frame.  $\left(\mathbf{R}_{wi_*}, \mathbf{t}_{wi_*}\right),  * \in \{ h, t \}$ denotes the rotation and translation of the host IMU frame and target IMU frame. $\mathbf{f}_i$ and $\mathbf{f}_t$ are calculated by $\mathbf{f}_i = \mathbf{R}_{ic} \mathbf{f}_c$ and $\mathbf{f}_t = \mathbf{R}_{c_tc_h} \mathbf{f}_c$, respectively. $\lfloor \cdot\rfloor_{\times}$ is the skew-symmetric operator.  $t_{cp}$ is obtained by $t_{cp} = \left(\mathbf{R}_{wc_h}^{\rm T}\mathbf{n}_w\right) \cdot \mathbf{f}_c$. And $d_c$ is the distance of plane in the host image frame, which is obtained by (\ref{fla: transformation matrix of plane}). 

The Jacobian of target IMU frame $\textbf{T}_{wi_t}$ is written as:
\begin{equation}
\begin{aligned}
\frac{\partial r_k}{\partial \delta \mathbf{T}_{wi_t}}= \mathbf{J}_I \cdot \frac{\partial \mathbf{p}'}{ \partial \Pi_c^{-1}(\mathbf{p}', d_p)} \cdot \frac{\partial \Pi_c^{-1}(\mathbf{p}', d_p)}{ \partial \delta \bm{\xi}_{wi_t}}
\end{aligned}
\end{equation}
where $\mathbf{J}_I$ and $\frac{\partial \mathbf{p}'}{ \partial \Pi_c^{-1}(\mathbf{p}', d_p)}$ are calculated in the same way as the Jacobian of the host IMU frame. $\frac{\partial \Pi_c^{-1}(\mathbf{p}', d_p)}{ \partial \delta \bm{\xi}_{wi_t}}$ is calculated by:
\begin{equation}
\begin{aligned}
\frac{\partial \Pi_c^{-1}(\mathbf{p}', d_p)}{ \partial \delta \bm{\xi}_{wi_t}}=
\left[\begin{matrix} 
\mathbf{J}_{rot} & \mathbf{J}_{trans}
\end{matrix}\right]
\end{aligned}
\end{equation}
where $\mathbf{J}_{rot}\in \mathbb{R}^{3\times3}$ is the Jacobian of the rotation part of the target IMU frame, and $\mathbf{J}_{trans}\in \mathbb{R}^{3\times3}$ is the Jacobian of the translation part of the target IMU frame. They are given by: 

\begin{equation}
\begin{aligned}
&\mathbf{J}_{rot} =  \mathbf{R}_{ic}^{\rm T} \lfloor\mathbf{R}_{wi_t}^{\rm T} \left(\frac{-\mathbf{R}_{wc_h}\mathbf{f}_c d_c}{t_{cp}} + \mathbf{t}_{wc_h} - \mathbf{t}_{wc_t}  \right)\rfloor_{\times} \\ 
&\mathbf{J}_{trans} =  -\mathbf{R}_{wc_t}^{\rm T}
\end{aligned}
\end{equation}

The Jacobian of the plane parameters is given by: 
\begin{equation}
\begin{aligned}
\frac{\partial r_k}{\partial \delta \left[\begin{matrix}\mathbf{n}_w \ d_w \end{matrix}\right]}= \mathbf{J}_I \cdot \frac{\partial \mathbf{p}'}{ \partial \Pi_c^{-1}(\mathbf{p}', d_p)} \cdot \frac{\partial \Pi_c^{-1}(\mathbf{p}', d_p)}{ \partial \delta \bm{\omega}_{\pi}}
\end{aligned}
\end{equation}
where $\mathbf{J}_I$ and $\frac{\partial \mathbf{p}'}{ \partial \Pi_c^{-1}(\mathbf{p}', d_p)}$ are calculated in the same way as the Jacobian of the host IMU frame. $\frac{\partial \Pi_c^{-1}(\mathbf{p}', d_p)}{ \partial \delta \bm{\omega}_{\pi}}$ is calculated by:

\begin{equation}
\begin{aligned}
\frac{\partial \Pi_c^{-1}(\mathbf{p}', d_p)}{ \partial \delta \bm{\omega}_{\pi}} = \left[\begin{matrix} \mathbf{J}_{n_w} \cdot \mathbf{J}_{n_gl} & \mathbf{J}_{d_w} \end{matrix}\right]
\end{aligned}
\end{equation}
where $\mathbf{J}_{n_w} \in \mathbb{R}^{3 \times 3}$ represents the Jacobian of the global parameters representation of the normal vector. $\mathbf{J}_{n_gl} \in \mathbb{R}^{3 \times 2}$ indicates the Jacobian of the global parameters with respect to the local parameters of the normal vector. $\mathbf{J}_{d_w} \in \mathbb{R}^{3 \times 1}$ denotes the Jacobian of the distance. $\mathbf{J}_{n_w}$ and $\mathbf{J}_{d_w}$ is written by: 
\begin{equation}
\begin{aligned}
\mathbf{J}_{n_w} &= -\frac{\mathbf{R}_{c_tc_h} \mathbf{f}_c \left(\mathbf{t}_{wc_h}^{\rm T} \cdot t_{cp}- \left(\mathbf{R}_{wc_h} \mathbf{f}_c\right)^{\rm T} \cdot d_c \right) }{t_{cp} \times t_{cp}} \\ 
\mathbf{J}_{d_w} &= -\frac{\mathbf{R}_{c_tc_h}\mathbf{f}_c}{t_{cp}}
\end{aligned}
\end{equation}
and $\mathbf{J}_{n_{gl}}$ is given by: 

\begin{equation}
\begin{aligned}
\mathbf{J}_{n_{gl}}=\left[\begin{array}{ccc}
-\cos(\psi)\sin(\phi) & -\sin(\psi) \cos(\phi) \\
\cos(\psi)\cos(\phi) & -\sin(\psi)\sin(\phi)  \\
0 & cos(\psi)  \\
\end{array}\right] 
\end{aligned}
\end{equation}
\bibliographystyle{IEEEtran}
\bibliography{mybibfile}

\end{document}